\documentclass[sigconf]{acmart}
\usepackage{booktabs} 
\usepackage{amsmath}
\usepackage{bm} 
\usepackage{tabularx}
\usepackage{caption}
\usepackage[normalem]{ulem}

 \copyrightyear{2018} 
 \acmYear{2018} 
 \setcopyright{acmcopyright}
 \acmConference[HRI '18]{2018 ACM/IEEE International Conference on Human-Robot Interaction}{March 5--8, 2018}{Chicago, IL, USA}
 \acmPrice{15.00}
 \acmDOI{10.1145/3171221.3171276}
 \acmISBN{978-1-4503-4953-6/18/03}

\usepackage[textfont=md,font=footnotesize]{caption}
\begin{document}
\title{Expressing Robot Incapability}

\author{Minae Kwon}
\affiliation{%
  \institution{Cornell University}
}
\email{mk898@cornell.edu}

\author{Sandy H. Huang}
\affiliation{%
  \institution{University of California, Berkeley}
}
\email{shhuang@cs.berkeley.edu}

\author{Anca D. Dragan}
\affiliation{%
  \institution{University of California, Berkeley}
}
\email{anca@cs.berkeley.edu}

\fancyhead{}

\newcommand{\eref}[1]{Eqn. (\ref{#1})}
\newcommand{\sref}[1]{Sec. \ref{#1}}
\newcommand{\figref}[1]{Fig. \ref{#1}}
\newcommand{\tabref}[1]{Table \ref{#1}}

\newcommand\Tstrut{\rule{0pt}{2.6ex}}         
\newcommand\Bstrut{\rule[-0.9ex]{0pt}{0pt}}   

\newcommand{\adnote}[1]{
 {\textcolor{blue}{\textbf{Anca: #1}}}}

\newcommand{\shnote}[1]{
 {\textcolor{red}{\textbf{SH: #1}}}}
 
\newcommand{\mknote}[1]{
 {\textcolor{purple}{\textbf{MK: #1}}}} 

\newcommand{\prg}[1]{\noindent\textbf{#1. }} 

\begin{abstract}
Our goal is to enable robots to express their incapability, and to do so in a way that communicates both \emph{what} they are trying to accomplish and \emph{why} they are unable to accomplish it. We frame this as a trajectory optimization problem: maximize the similarity between the motion expressing incapability and what would amount to successful task execution, while obeying the physical limits of the robot. We introduce and evaluate candidate similarity measures, and show that one in particular generalizes to a range of tasks, while producing expressive motions that are tailored to each task. Our user study supports that our approach automatically generates motions expressing incapability that communicate both \emph{what} and \emph{why} to end-users, and improve their overall perception of the robot and willingness to collaborate with it in the future.
\end{abstract}

\maketitle

\section{Introduction}
As robots become increasingly capable, they may unintentionally mislead humans to overestimate their capabilities ~\cite{Cha_2015}. Thus, it is important for a robot to communicate when it is incapable of accomplishing a task. There are two relevant pieces of information when expressing incapability: \emph{what} the task is, and \emph{why} the robot is incapable of accomplishing it.

Understanding \emph{why} the robot is incapable gives observers a better understanding of its capabilities, which improves joint human-robot task performance~\cite{Nikolaidis_2017}. Transparency about the causes of incapability also helps observers assign blame more accurately~\cite{Kim_2006}. If observers also understand \emph{what} the robot was trying to do, they are better able to help the robot complete the task~\cite{Nicolescu_2001,Tellex_2014,Hayes_2016}.

One of the simplest ways to express incapability is to carry out the failure. Unfortunately, not all failures are inherently communicative about the \emph{what} and the \emph{why}. The fact that the robot failed to complete the task means that it might not have gotten far enough in the task for the \emph{what} to become obvious---in fact, robots sometimes fail before they even start. Our goal in this work is to expressively show a robot's incapability, beyond simply failing.

\begin{figure}[t!]
\centering
\includegraphics[width=0.8\columnwidth]{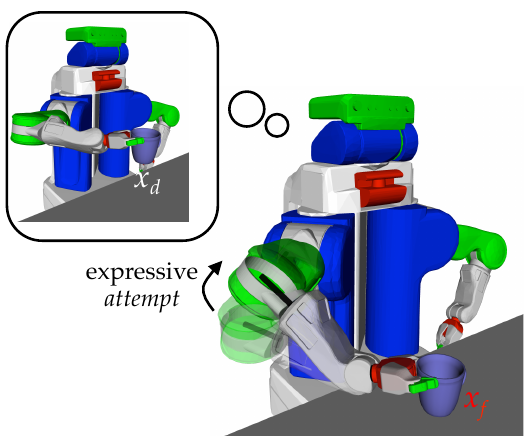}
\caption{We introduce a method to generate motion for incompletable tasks that communicates both the intended goal of the task and why the robot is incapable of completing the task. The method generates an \emph{attempt} motion meant to resemble successful execution (e.g., moving the end-effector from $x_f$ to $x_d$) while obeying the constraints on the robot's limitations. In this example, the robot ends up lifting its elbow to communicate that it is trying to lift the cup, but the cup is too heavy for it.}
\label{fig:front_fig}
\end{figure}

In general, even just acknowledging an incapability (e.g., via language or motion) mitigates the damage to human perception of the robot~\cite{Lee_2010,Takayama_2011,Brooks_2016}; in some situations, demonstrating an incapability may actually increase the likeability of the robot~\cite{Mirnig_2017,Ragni_2016}, due to the Pratfall Effect~\cite{Aronson_1966}. If we can further ensure better understanding, we hope that people will not only evaluate the robot more favorably, but they will also be able to make more accurate generalizations of the robot's capability across different tasks.

We focus on the robot's motion as the communication channel, which has already been established as an effective and natural way of communicating a robot's intent~\cite{Dragan_2013}.
\begin{quote}
\emph{Our key insight is that a robot can express both \emph{what} it wants to do and \emph{why} it is incapable of doing it by solving an optimization problem: that of executing a trajectory similar to the trajectory it would have executed had it been capable, subject to constraints capturing the robot's limitations.}
\end{quote}

Take lifting a cup that is too heavy as an example, or turning a valve that is stuck. Once the robot realizes that it is incapable of completing the task, the robot would find some motion that still conveys what the task is and sheds light on the cause of incapability, all without actually making any more progress on lifting the cup or turning the valve. We call this motion an \emph{attempt}; \figref{fig:front_fig} shows what an attempt might look like for the lifting example. 

We focus on tasks like these, in which the robot is unable to move its end-effector (and the object it is manipulating) to the desired goal pose. Although this is a relatively narrow set of tasks that the robot may be incapable of completing, it is a step toward \emph{automatically} generating task-specific motions expressing incapability: prior work relied on either simple strategies~\cite{Kobayashi_2005,Nicolescu_2001} or motions hand-crafted for each specific situation in which the robot is incapable~\cite{Takayama_2011}.

Our main contribution is to frame the construction of expressive incapability trajectories as a trajectory optimization problem. We explore several reasonable objectives for this optimization problem, and find that one generalizes best across a range of incompletable tasks. Our user study shows that attempt trajectories significantly improve not only participants' understanding of what the robot is trying to do and why it cannot, but also their overall perception of the robot and willingness to collaborate with it in the future.
\section{Related Work}
Our motions communicate both the what and the why, i.e., both intent and the cause of incapability.

\noindent\textbf{What/Intent.} Much work has focused on motion for conveying robot intent, i.e., what task the robot is doing \cite{Gielniak_2011,Szafir_2014,Dragan_2013}. What is different in our work is that the robot is incapable of actually doing the intended task, so it needs a way to convey enough about the task without being able to actually do it. This is our idea of \emph{attempt}: in our work, the robot generates a (failed but expressive) attempt at the task. We focus on how to autonomously generate such attempts.

Communicating intent is useful: legible motion improves joint human-robot task performance~\cite{Dragan_2015} and in fact arises naturally from optimizing for joint performance~\cite{Stulp_2015}. Beyond motion, prior work has explored communicating intent through visualizing planned trajectories (e.g., via targeted lighting~\cite{Augustsson_NordiCHI2014} or augmented reality~\cite{Carff_2009,Rosen_2017}), gaze~\cite{Mutlu_2009}, body language~\cite{Breazeal_2000}, human-like gestures~\cite{Haddadi_2013,Gleeson_2013}, verbal communication ~\cite{raman2013sorry}, and LED displays~\cite{Matthews_2017}. 

\noindent\textbf{Why/Cause of Incapability.} Prior work on using motion to communicate \emph{why} a robot cannot complete a task relied on simple strategies: moving back-and-forth when stuck in front of an obstacle~\cite{Kobayashi_2005}, or repeatedly executing a failing action~\cite{Nicolescu_2001}. In the context of lifting a cup that is too heavy, the latter approach would result in a trajectory that repeatedly reaches for the cup, grasps it, then rewinds. We show in ~\sref{sec:mainstudy} that our approach significantly improves identification of the task goal and cause of incapability, compared to the latter method. Another approach relies on hand-designed motions, crafted per-task using animation principles, to indicate recognition of success or failure in completing the task~\cite{Takayama_2011}. In contrast, our approach of optimizing for motions expressing incapability generalizes to multiple tasks, while resulting in an attempt trajectory tailored to each task.

Communicating why a robot is incapable is closely related to work that examines how robots can warn before failing. Robots can forewarn users of possible failures through text~\cite{Lee_2010} or confidence levels~\cite{Desai_2013,Kaniarasu_2013}, trajectory timings~\cite{Zhou_2017}, and actively choosing actions that showcase failure modes~\cite{Nikolaidis_2017}. Setting accurate expectations of robot capabilities is important for narrowing the gap between the perceived and true capabilities of the robot ~\cite{Cha_2015}.
\section{Expressing Incapability, Formalized}
\prg{Notation}
A robot's trajectory $\xi$ is a sequence of $T$ robot configurations: $\xi_t$ is the configuration of the robot at time $t$. $\phi_b: \mathcal{Q} \mapsto \text{SE}(3)$ is the forward kinematics function at body point $b$, and thus gives the pose (rotation and translation) of the body part at that point. $\phi_b': \mathcal{Q} \mapsto \mathbb{R}^3$ 
gives the translation of body point $b$. The body points we consider in our particular implementation are $ee$ (end-effector), $el$ (elbow), $sh$ (shoulder), and $ba$ (base), which we found to be a reasonable discretization of the arm.

\prg{Incompletable Task Definition}
A task is defined by a starting configuration $q_s$, along with a desired final pose $x_d$ for the end-effector (or, in more specific instances, a desired configuration $q_d$). \figref{fig:front_fig} shows an example of $x_d$ for the task of lifting a cup. There may also be additional constraints that define the task, such as the need to keep contact with an object as the robot moves its end-effector from $\phi_{\text{ee}}(q_s)$ to $x_d$.

An \emph{incompletable} task is one in which the end-effector cannot make progress beyond a certain failure point $x_f$. For instance, if the cup the robot tries to lift is too heavy, then $x_f$ would be the pose of the end-effector when it first grasps the cup (because it can no longer proceed in the task from there due to the cup's weight). $x_d$ would be vertically above $x_f$: the location to which, if the robot were holding on to the cup, the cup would be lifted. \figref{fig:front_fig} also shows $x_f$ for this example.

\prg{Expressing Incapability as an Optimization Problem}
The goal of expressive incapability trajectories is to communicate \emph{what} the robot was attempting to do and \emph{why} it is incapable of it. A simple approach would be to move to $x_f$ (i.e., as far as we could get with the task), and stop. Prior work has suggested also repeating this motion \cite{Kobayashi_2005,Nicolescu_2001}. We hypothesize we can do better.

Our idea is to continue the task by executing an \emph{attempt} trajectory past the failure point. Our insight is that we can formalize this as an optimization problem: find an attempt trajectory that maximizes similarity to the trajectory from $x_f$ to $x_d$ that would have been executed, had the incapability not existed. We solve this optimization subject to the incapability constraint that the end-effector cannot proceed further. 

We capture similarity, or rather dissimilarity, via a cost function $c(\xi,x_f,x_d)$, and find the attempt trajectory as:
\begin{equation}
\begin{aligned}
\xi^* =\  & \underset{\xi}{\text{argmin}}
& &  c(\xi; x_f, x_d) +\frac{1}{\lambda} \sum_{t=0}^{T-1} \|\xi_{t+1}-\xi_t\|^2\\
& \text{subject to}
& & \phi_{ee}(\xi_t) = x_f, \; \forall t \in \{0..T\}\\
&&& \text{collision-free}(\xi).
\end{aligned}
\label{eqn:prob_form}
\end{equation}
This objective trades off between the similarity cost and a smoothness term common in trajectory optimization \cite{Zucker_IJRR2013, Schulman_IJRR2014}. 

\prg{Cost Functions}
Crucial to generating a good attempt trajectory is finding a good cost function $c(\xi;x_f,x_d)$. We investigate cost functions that seek to mimic the change from $x_f$ to $x_d$. But since the end-effector cannot move in $\xi$, $c$ cannot just consider the end-effector's motion: it has to consider the configuration space.

If the desired configuration $q_d$ is not provided, we define it as the inverse kinematics solution for $x_d$ that is closest to the starting configuration $\xi_0$ for the attempt:
\begin{equation}
\begin{aligned}
& q_d =
& & \underset{q}{\text{argmin}}
&&& \|q - \xi_0\|^2\\
& \text{ }
& & \text{subject to}
&&& \phi_{\text{ee}}(q) = x_d.
\end{aligned}
\end{equation}

\noindent\emph{\textbf{Configuration-Based Cost $ c_\text{q}$: }}A natural starting point is to try to mimic in $\xi$ the change in \emph{configuration} from $\xi_0$ (with the end-effector at the failure point $x_f$) to $q_d$:
\begin{equation}
        c_\text{q}(\xi; x_f, x_d) = d(\xi_T - \xi_0, q_d - \xi_0),
    \end{equation}
where $d$ is some distance metric such as the $\ell_2$-norm (we discuss options below).

\noindent\emph{\textbf{Workspace-Based Cost $ c_\text{b}$: }}Since configuration spaces can sometimes be counterintuitive, we also look at a cost that tries to mimic, for each body point, the change in \emph{position} for that body point:
\begin{equation}
        c_\text{b}(\xi; x_f, x_d) = \sum_{b \in B} d(\phi_b'(\xi_T) - \phi_b'(\xi_0), \phi_b'(q_d) - \phi_b'(\xi_0))
    \end{equation}
Despite the end-effector staying put, this incentivizes, for instance, the elbow to move in the same direction it would have moved had the task been successful.    

\noindent\emph{\textbf{Emulate End-Effector Cost $ c_\text{ee}$: }}We also introduce a third, somewhat less obvious cost function. Since the end-effector is central to the task, and now it cannot proceed further, this cost function tries to mimic using the \emph{other} body points what the end-effector would have done:
\begin{equation}
        c_\text{ee}(\xi; x_f, x_d) = \sum_{b \in B} d(\phi_b'(\xi_T) - \phi_b'(\xi_0), x_d' - x_f')
    \end{equation}

\prg{Distance Metrics}
Each of these costs relies on a distance metric between vectors. We consider three distance metrics $d(v_1,v_2)$:
\begin{enumerate}
    \item The squared $\ell_2$-norm encourages $v_1$ and $v_2$ to have similar direction and magnitude:
    \begin{equation}
        d_{\ell\text{2}}(v_1,v_2) = \|v_1 - v_2\|^2.
    \end{equation}
    \item The (negative) dot product encourages $v_1$ and $v_2$ to have similar direction and large magnitudes:
    \begin{equation}
        d_\text{dot}(v_1,v_2) = - \, v_1 \cdot v_2 = - \, \|v_1\| \, \|v_2\| \cos\theta.
    \end{equation}
    \item We also introduce a generalization of the dot product that uses a hyperparameter $k$ to control the trade-off between $v_1$ and $v_2$ having similar direction versus large magnitudes:
    \begin{equation}
    \begin{aligned}
        & d_\text{proj}(v_1,v_2; k)
        & & = - \, v_1 \cdot v_2 \, \bigg(\frac{v_1 \cdot v_2}{\| v_1 \| \, \| v_2 \| }\bigg)^{k-1} \\
        &&& = - \, \| v_1 \| \, \| v_2 \| \, (\cos \theta)^k.
    \end{aligned}
    \end{equation}
\end{enumerate}

The last two distance metrics are motivated by the fact that a larger magnitude makes the attempt trajectory more obvious to human observers. Intuitively, $d_\text{proj}$ projects $v_1$ onto $v_2$, projects the result back onto $v_1$, projects the result onto $v_2$, and so on. The hyperparameter $k$ defines how many times this projection happens, so for larger $k$, matching direction matters more than large magnitudes. Note that $d_\text{proj}(v_1,v_2;1)$ = $d_\text{dot}(v_1,v_2)$.

\begin{figure}[t!]
\centering
\includegraphics[width=\columnwidth]{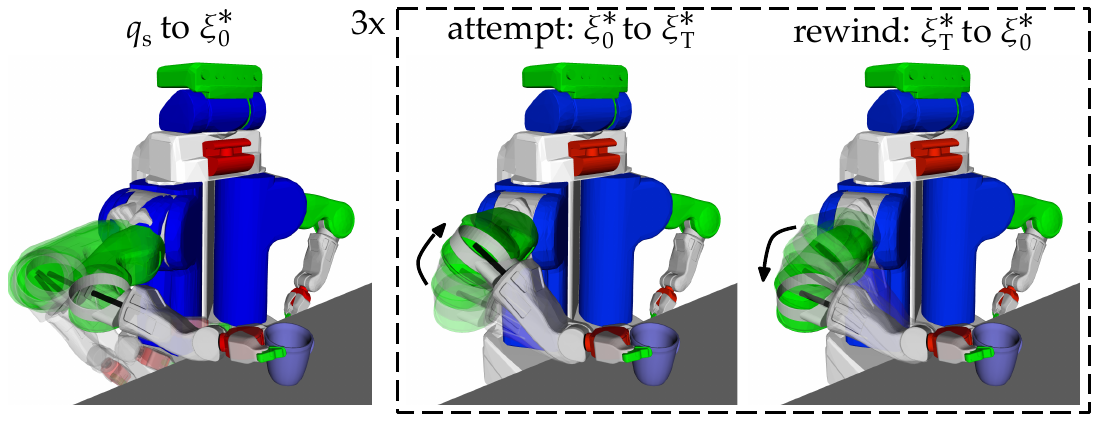}
\caption{For a given incompletable task, the robot first executes the task until the point of failure (left), at which point it executes the attempt trajectory $\xi^*$ (center). To emphasize this motion, the robot then executes the reverse of $\xi^*$ to rewind back to $\xi^*_0$ (right), and repeats this two more times.}
\label{fig:approach}
\end{figure}

\prg{Overall Attempt}
The robot starts at $q_s$, and moves along the normal task execution trajectory to the point of failure $x_f$; at this point its configuration is $\xi^*_0$, where $\xi^*$ is the optimum from \eref{eqn:prob_form}. From there, it executes $\xi^*$.

Since prior work on using motion to express incapability found repetitions to be useful~\cite{Kobayashi_2005,Nicolescu_2001}, we also explore rewinding and repeating: the robot executes the reverse of $\xi^*$ to get back to $\xi^*_0$, and repeats the execute-rewind twice more, as in \figref{fig:approach}.

In what follows, we first show the outcome $\xi^*$ that each cost function leads to, and use this to select a good cost function. We then run experiments to determine the appropriate relative timing of the $\xi^*$ and the rewind (to enhance expressiveness~\cite{Zhou_2017}), as well as whether repetitions of the attempt help. Armed with the right general parameters, we conduct a main study across different incompletable tasks to test whether these motions, optimized to be more expressive, lead to better understanding of what task the robot is trying to do and why it will fail.

\section{Comparing Cost Functions}
In this section, we contrast the different behaviors produced by the cost functions and distance metrics.

\begin{figure}[t!]
\centering
\includegraphics[width=\columnwidth]{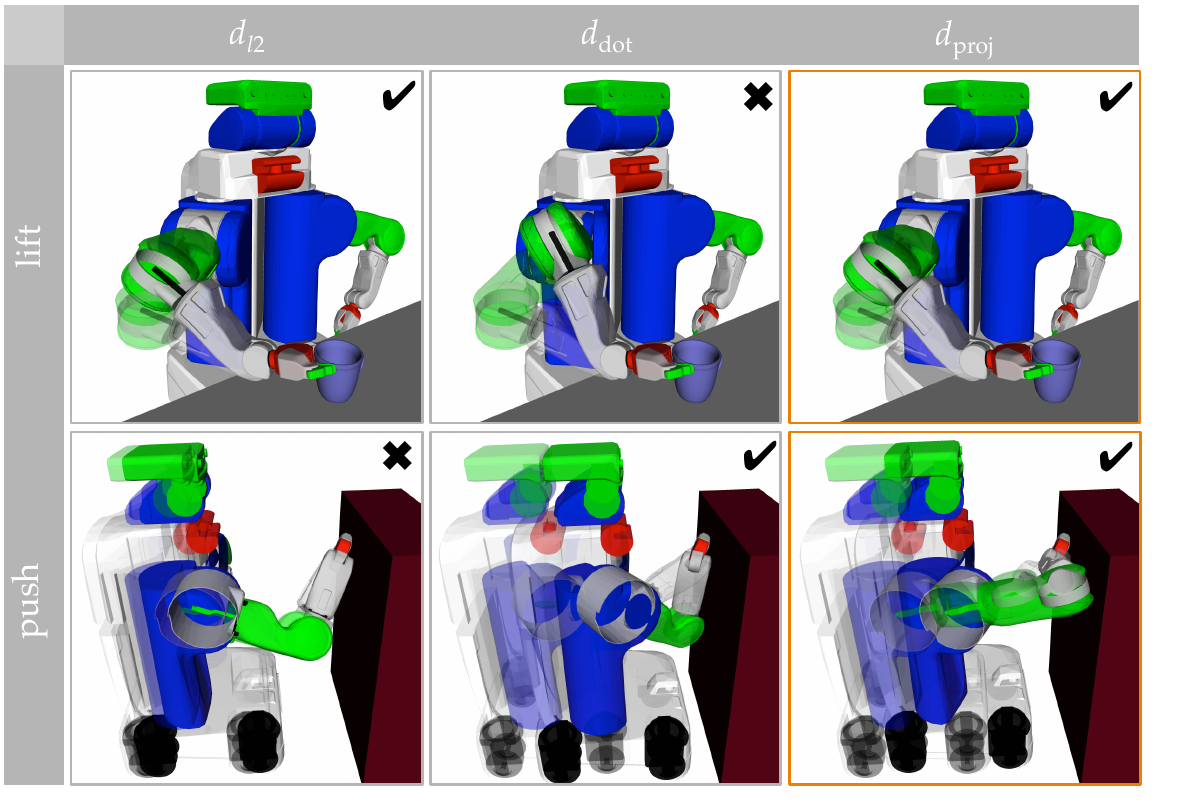}
\caption{Attempt trajectories $\xi^*$ that optimize cost function $c_\text{ee}$ with each of the three proposed distance metrics. Each image shows $\xi_0$ (transparent) and $\xi_T$ for that attempt trajectory. $d_\text{proj}$ (last row) results in communicative attempt trajectories for both the \emph{lift} and \emph{push} tasks.}
\label{fig:compare_dists}
\end{figure}

\begin{figure*}[t!]
\centering
\includegraphics[width=0.75\textwidth]{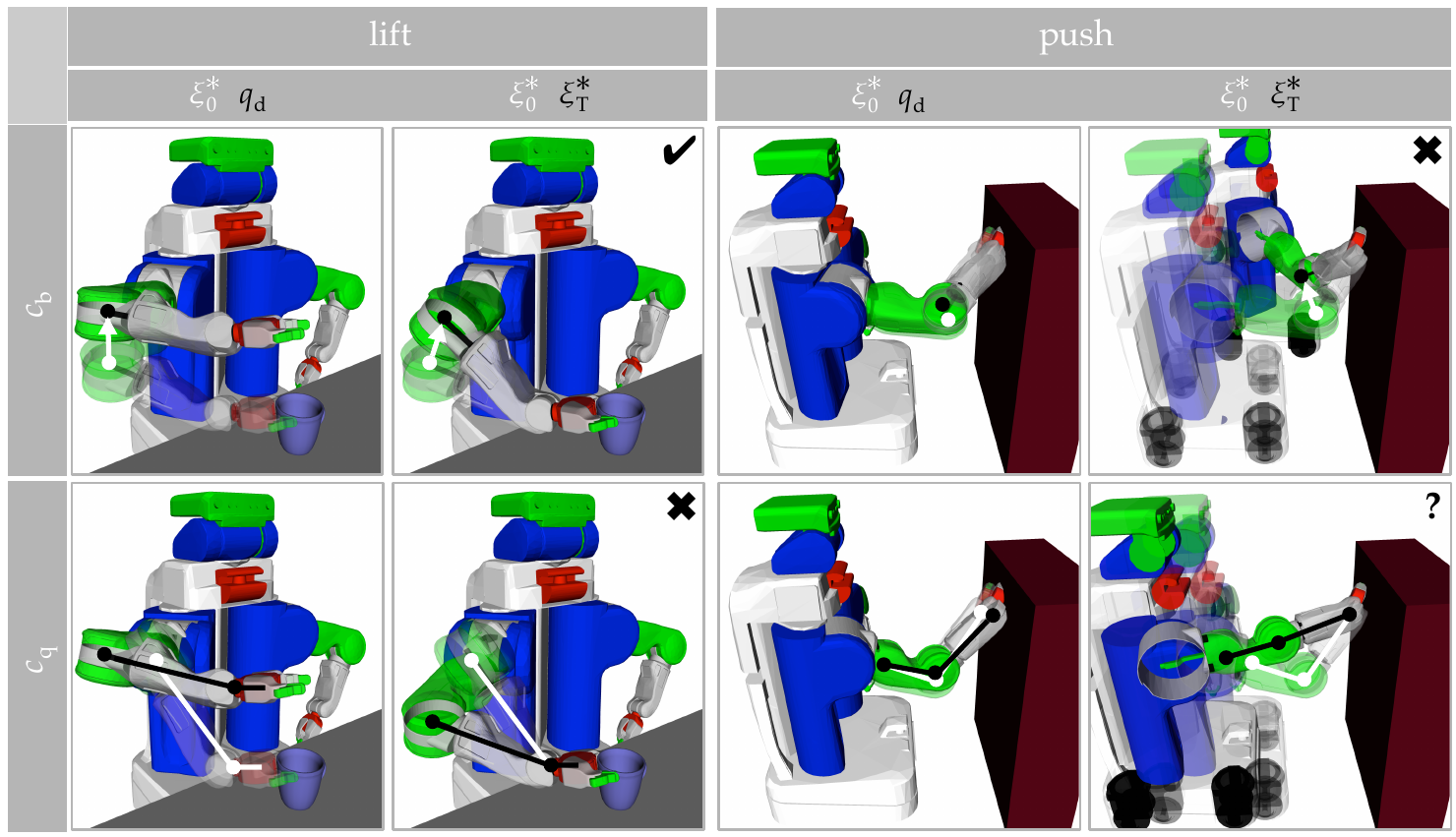}
\caption{Attempt trajectories $\xi^*$ that optimize cost function $c_\text{b}$ or $c_\text{q}$, with distance metric $d_\text{proj}$. When optimizing for $c_\text{b}$, the attempt trajectory for \emph{lift} is communicative, but for \emph{push} the robot swings out to the left, which does not indicate that it is trying to push. When optimizing for $c_\text{q}$, the attempt trajectory for \emph{push} is reasonable (although it could be confused for pulling, since the robot moves away from the shelf), but for \emph{lift} the robot's elbow moves downward, which does not indicate that it is trying to lift.}
\label{fig:compare_costs_cb_cq}
\end{figure*}

\begin{figure*}[t!]
\centering
\includegraphics[width=\textwidth]{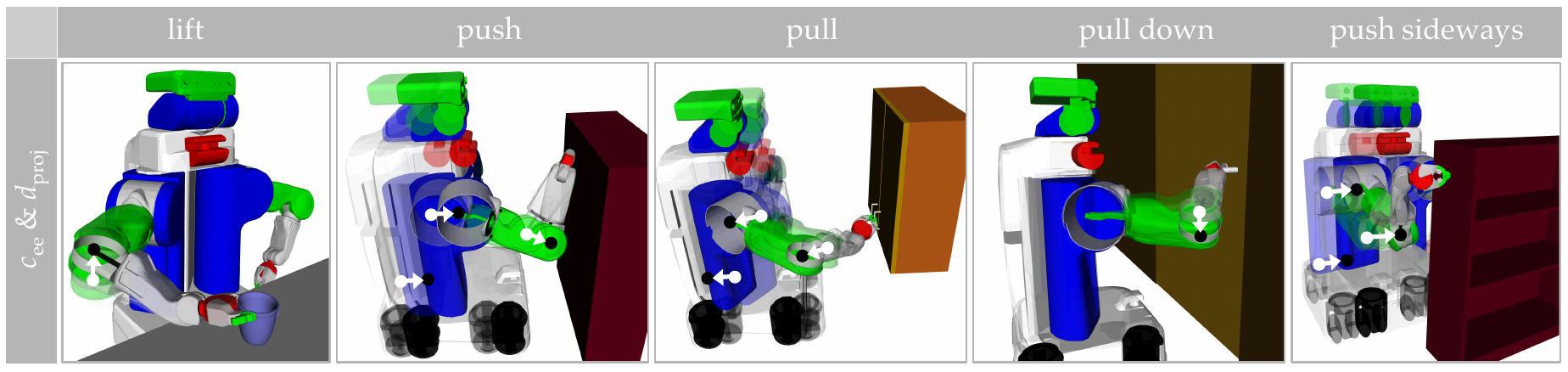}
\caption{Attempt trajectories $\xi^*$ that optimize cost function $c_\text{ee}$ with distance metric $d_\text{proj}$, for five incompletable tasks. Arrows show the direction of movement for the considered body points (elbow, shoulder, and base)---for each task, these body points imitate how the robot's end-effector would move, if it were able to successfully accomplish the task.}
\label{fig:all_tasks}
\end{figure*}

\subsection{Implementation Details}
We use a simulated PR2 robot in OpenRAVE~\cite{Diankov_2010} and optimize for attempt trajectories using TrajOpt~\cite{Schulman_IJRR2014}. Since costs $c_\text{b}$ and $c_\text{q}$ depend on $q_d$, which in turn depends on $\xi_0$, we simplify the optimization problem by optimizing over $\xi_{1:T}$ for each possible starting configuration $\xi_0$ of the attempt trajectory (found by running an inverse kinematics solver on $x_f$), and then select the full trajectory $\xi$ that minimizes the objective function.

We use grid search to select the hyperparameter $\lambda$ and a bias $\alpha$ separately for each cost function and distance metric pair.\footnote{We found that optimization is more stable if the terms in the objective are in the same range. So, we add a bias $\alpha$ to the cost function: $c'(\xi;x_f,x_d) = c(\xi;x_f,x_d) + \alpha$. We select $\lambda \in \{10,20,40,80,160\}$ and $\alpha \in \{0,0.3,0.6,1.0,2.0\}$.} For $c_\text{b}$, $B = \{ \text{el}, \text{sh} \}$ and for $c_\text{ee}$, $B = \{ \text{ba}, \text{el}, \text{sh} \}$.\footnote{This is because matching the configuration would not make as much sense for the base, but the base can be useful as another body point with which to mimic the end-effector motion. In general, which body points to use might be a robot-specific question, to be determined from a few tasks and generalized to new tasks.} We chose $k = 9$ for $d_\text{proj}$.

\subsection{Behaviors} Overall, we found that $c_\text{ee}$ (the cost that mimics the desired end-effector motion with the other body points) with $d_\text{proj}$ (the distance metric that generalizes the dot product) is a combination that reliably leads to attempt trajectories that both move in a way that makes the task clear, and have enough movement to be noticeable. We explain this finding below by first contrasting distance metrics, and then contrasting cost functions. We use two incompletable tasks for this contrast: lifting a cup that is too heavy (the \emph{lift} task), and pushing a shelf that is immovable (the \emph{push} task).

\prg{Explanation of Attempt Behavior}
\figref{fig:compare_dists} shows the results of $c_\text{ee}$ with each distance metric. Across the board for \emph{lift}, optimizing for $c_\text{ee}$ encourages the robot to use its elbow to produce the motion that the end-effector would otherwise produce. We thus see the robot \emph{lifting} its elbow while keeping the end-effector on the cup that is too heavy to lift. Across the board for \emph{push}, the robot is using its elbow, shoulder, and base, to mimic the end-effector forward motion. As a result, the robot moves \emph{forward} toward the shelf, as the end effector stays put, unable to actually push the shelf.

\prg{Distance Comparison}
$d_\text{proj}$ works across both tasks. In contrast, using the $d_{\ell\text{2}}$ distance metric results in an attempt trajectory for \emph{push} that barely moves, and using the $d_\text{dot}$ distance metric results in an over-exaggerated motion for \emph{lift} in which the robot's elbow twists toward the center.

\prg{Cost Comparison}
Now we turn to examining the performance of the other two cost functions ($c_\text{b}$ and $c_\text{q}$) with the best distance metric $d_\text{proj}$. 

Optimizing for $c_\text{b}$ results in a confusing attempt trajectory for \emph{push} where the robot swings to the left. This is because the elbow and shoulder body points move slightly to the left from $\xi_0$ to $q_d$: if the robot were successful in pushing, its end-effector would move further out, extending the arm, and the elbow would no longer protrude to the right, and instead move inward (to the left of the robot). The optimization thus selects an attempt trajectory that moves the elbow and shoulder as far as possible inward along this general direction (\figref{fig:compare_costs_cb_cq}). We observe that moving the other body points (e.g., the elbow or shoulder) in the way they ideally would during a successful task execution is not always indicative of the task.

Optimizing for $c_\text{q}$ results in a confusing attempt trajectory for \emph{lift}, where the robot's elbow moves downward to match the desired configuration $q_d$ (\figref{fig:compare_costs_cb_cq}). In the attempt for \emph{push}, the robot moves away from the shelf rather than toward it, also to match $q_d$---which would have the arm extended out after a successful push. This could work, but could also be mistaken for pulling instead of pushing. We observe that because configuration spaces are often counterintuitive, mimicking the motion in configuration space can lead to surprising, counterintuitive motions.

In contrast, it seems that using the other body points to imitate what the end-effector cannot do might actually be indicative of what the robot is trying to achieve. We put this to the test in \sref{sec:mainstudy}. But first, we tune the hyperparameters of attempts---the timing of the motions, and whether to include repetitions.

\prg{$c_{\text{ee}}$ Across More Tasks} Optimizing for $c_\text{ee}$ with $d_\text{proj}$ also generates communicative attempt trajectories for other incompletable tasks, shown in \figref{fig:all_tasks}: opening a locked cabinet (the \emph{pull} task), turning a locked door handle (the \emph{pull down} task), and pushing a shelf to the side (the \emph{push sideways} task). Across all five tasks, we set $k=3$, $\lambda = 20$, and $\alpha = 0.3$.\footnote{To simplify optimization, we additionally assume a fixed base when computing forward kinematics for non-base body points.} These are the attempt trajectories that we show in our user studies (in video form). A video summary of the cost comparisons and attempt motions for each task is at \href{https://www.youtube.com/watch?v=uSnUtpcdlck}{youtu.be/uSnUtpcdlck}.

\section{Timing Motions That Express Incapability}
Our aim was to manipulate timing in order to enhance the expressiveness of our optimized motions. We temporally divided a motion expressing incapability into attempt and rewind motions. The attempt motion consists of the trajectory $\xi^*$ produced by the cost function optimization from \eref{eqn:prob_form}. The rewind motion, which is the reverse of the attempt trajectory, immediately follows the attempt motion. Our goal in this study was to find the pair of timings for the attempt and rewind motions that best convey a robot's task goal and the cause of incapability. 

\subsection{Experiment Design}
\prg{Manipulated Variable}
We manipulated timing in this study and chose three speeds (Fast, Moderate, and Slow). We were only interested in the \emph{relative} speed between the attempt and rewind motions, so we fixed the rewind speed at Moderate and varied the attempt speed, creating three conditions: Fast attempt with Moderate rewind (Fast, Moderate), Slow attempt with Moderate rewind (Slow, Moderate), and Moderate attempt with Moderate rewind (Moderate, Moderate).

\prg{Other Variables}
We tested timing across the five tasks in \figref{fig:all_tasks}.

\prg{Subject Allocation}
We recruited 60 participants (37\% female, median age 32.5) via Amazon Mechanical Turk (AMT). All participants were from the United States and had a minimum approval rating of 95\%. Timing was within-subjects: participants saw each of the three timing conditions. Task type was between-subjects: participants saw only one type of task. 

\begin{figure}[t!]
\centering
\hbox{\hspace{-0.5em}\includegraphics[width=0.5\textwidth]{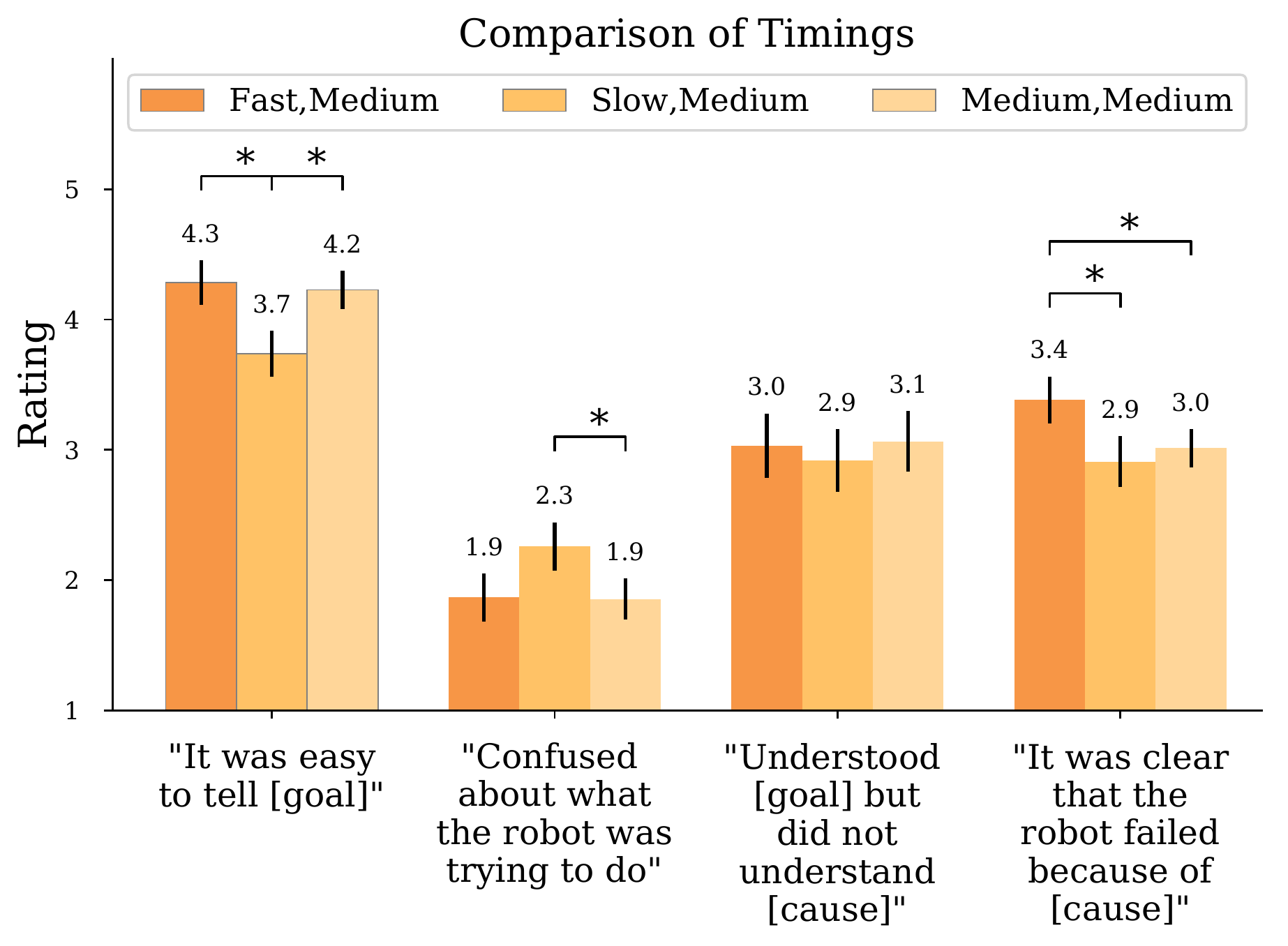}}
\caption{
Average ratings toward different timings. Timing was within-subjects, meaning participants rated each of the timing pairs. Overall, participants preferred Fast attempt and Moderate reset.}
\label{fig:timing}
\end{figure}

\prg{Dependent Variables}
\label{sec:depvar}
Participants saw videos of all three timings. We explained to the participants what the robot was trying to do (its intended goal), and why it could not complete the task. We then asked them to help us select the timing that best expresses both the goal and cause of incapability. We created four statements to assess each timing (\figref{fig:timing}), and asked participants to rate their level of agreement with these statements on a 5-point Likert scale. We also asked participants to rank the three timings.

\subsection{Analysis}
We ran a repeated-measures ANOVA with timing as a factor and user ID as a random effect, for each item. We found significant effects of timing on how easy it was to tell the goal ($F(2,118)=8.26$, $p=.0004$), how confusing the goal was ($F(2,118)=4.47$, $p=.013$), and how clear the cause was ($F(2,118)=5.13$, $p=.007$). 
Across the board, the timing that worked best was (Fast, Moderate), as shown in \figref{fig:timing}. 58\% of participants ranked this timing first. This indicates the attempt part of the motion should be faster than the rewind, which intuitively makes sense---it perhaps conveys that the attempt portion is the purposeful action on which the robot expends more energy, and the rest is at a normal speed that the robot would use to move around. We use this (Fast, Moderate) timing in our main study, described in \sref{sec:mainstudy}.

\section{Comparing Repeated and Non-Repeated Attempts}
After determining the best timing for the attempt and rewind motions, we looked at whether including repetition for the attempt motions enhances expressiveness. 
\subsection{Experiment Design}
\prg{Manipulated Variable}
We manipulated repetition, where we compared N=3 iterations of the attempt motion with a single (N=1) iteration of the attempt motion. 

\prg{Dependent Variables}
We used the same measures as in \sref{sec:depvar}.

\prg{Subject Allocation}
We recruited 60 participants (47\% female, median age 35) via AMT. All participants were from the United States and had a minimum approval rating of 95\%. Repetition type was within-subjects: every participant saw N=1 and N=3 iterations of the attempt motion. Task type was between-subjects: participants saw only one type of task. 

\subsection{Analysis}
\begin{figure}[t!]
\centering
\hbox{\hspace{-0.5em}\includegraphics[width=0.5\textwidth]{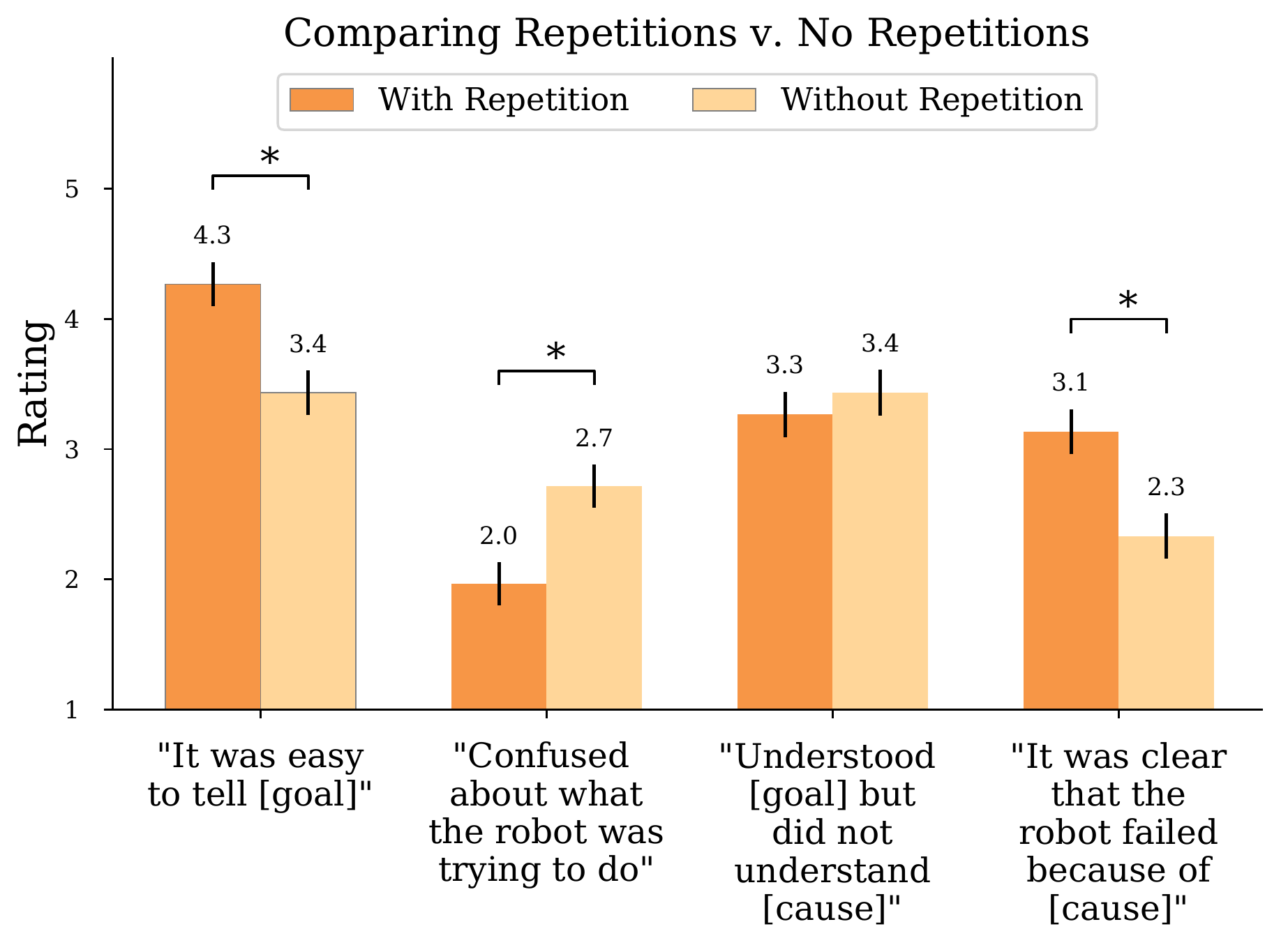}}
\caption{
Average ratings comparing repetitions with no repetitions. The study was within-subjects, meaning participants rated both motions with repetition and motions without repetition. Overall, participants preferred motions with repetitions.}
\label{fig:repetition}
\end{figure}

We ran a repeated-measures ANOVA with timing as a factor and user ID as a random effect, for each item. We found repetitions significantly increased how easy it was to tell the goal ($F(1,58.14)=20.21$, $p<.0001$), decreased confusion about the goal ($F(1,62.71)=15.95$, $p=.0002$), and made the cause more clear ($F(1,63.64)=16.94$, $p=.0001$). \figref{fig:repetition} shows the results. We proceed with repetitions for our main study.

\section{Main Study: Is Expressive Motion Expressive?}

\label{sec:mainstudy}
With the details of how to generate motions expressing incapability out of the way, we now turn to how much this helps people identify the robot's goal and cause of incapability.

\subsection{Experiment Design}
\prg{Manipulated Variable}
We compared our motions expressing incapability against the state-of-the-art approach for automatically generating motion to express robot incapability \cite{Kobayashi_2005,Nicolescu_2001}. With the state-of-the-art approach, the robot repeatedly executes a failing action. The manipulated variable was generating the attempt motion via our optimization-based approach versus via the repeated-failures approach.

We created motions expressing incapability following the process in \figref{fig:approach}. For each task, we optimize $c_\text{ee}$ with $d_\text{proj}$ to generate the attempt $\xi^*$. The robot moves from $q_s$ to $\xi_0^*$, executes the attempt at speed=Fast, rewinds back to $\xi_0^*$ at speed=Moderate, and repeats the attempt-rewind two more times for a total of N=3 iterations.

For the repeated-failure motions, the robot moves from $q_s$ to $\xi_0^*$, rewinds back $T$ time steps at speed=Moderate, and moves back to $\xi_0^*$ at speed=Fast. The robot rewinds $T$ time steps and moves back to $\xi_0^*$ two more times for a total of N=3 iterations. The timing and number of iterations are the same as for our approach, to limit possible confounds.

\begin{center}
\begin{tabular}{c l} 
\hline
 \multicolumn{2}{ c }{Incorrect Goal Recognition Statements} \Tstrut\Bstrut\\
\hline
Lift & The robot was trying to push the cup. \Tstrut \\
     & The robot was trying to pull the cup.\\
     & The robot was trying to knock over the cup. \Bstrut \\
\hline
Pull & The robot was trying to slide the cabinet door \Tstrut \\
     & \hspace{2mm} sideways.\\
     & The robot was trying to sense the cabinet handle. \\
     & The robot was trying to push the cabinet away. \Bstrut \\
\hline
Pull  & The robot was trying to sense the door handle. \Tstrut \\
Down  & The robot was trying to prevent someone from\\
      & \hspace{2mm} opening the door on the other side. \\
      & The robot was trying to remove the door handle. \Bstrut \\
\hline 
Push & The robot was trying to sense the box. \Tstrut \\
     & The robot was trying to stroke the box. \\
     & The robot was trying to knock on the box. \Bstrut \\
\hline
Push     & The robot was trying to sense the shelf. \Tstrut \\
Sideways & The robot was trying to lift the shelf. \\
         & The robot was trying to knock on the shelf. \Bstrut \\
\hline
\end{tabular}
\end{center}
\vspace{1 mm}
\begin{center}
\begin{tabularx}{\linewidth}{X}
  \hline
  \multicolumn{1}{c}{Incorrect Cause of Incapability Statements} \Tstrut \Bstrut \\
  \hline
     The robot had a mechanical failure (e.g. ran out of battery, arm  \Tstrut \\
     \hspace{2mm} got stuck, etc.) or software crash.  \\
     The robot did not know how to [goal].\\
     The robot is waiting for permission to [goal]. \Bstrut \\
  \hline
\end{tabularx}
\hfill \break
\vspace{2 mm}
\captionsetup{type=table}
\caption{List of incorrect plausible goal and cause of incapability statements participants had to choose from. The cause of incapability statements were similar across tasks.}
\label{table:incorrect_goal+cause}
\end{center}

\prg{Dependent Variables}
Our dependent variables included how well participants could infer the robot's goal and cause of incapability, as well as measures regarding their perception of the robot.

We assessed \emph{\textbf{goal recognition}}---how well participants could infer the intended task goal---in several ways. First, using an open-ended response, we asked participants to state what they thought the task goal was. Second, we presented four plausible task goals, with the correct task goal as once of the choices. We then asked participants to rate, on a 5-point Likert scale labeled ``Strongly Disagree'' to ``Strongly Agree,'' how well each task goal described what the robot's goal was. Lastly, we asked participants to explicitly rank the task goals in order of how well they described the robot's goal. Incorrect goal alternatives are described in \tabref{table:incorrect_goal+cause}. 

We measured \emph{\textbf{cause of incapability recognition}}---how well participants infer the incapability underlying the robot's failure---in a similar way. The only difference was that the Likert-scale questions included a second correct option (``The robot was not strong enough [to complete the task]'') because it is a plausible interpretation of our motion: our method is not meant to differentiate between, for instance, the cup being too heavy and the robot not being strong enough. Rather, our method is meant to differentiate between these two correct options and other causes of incapability that are possible, but untrue, for instance the robot running out of battery, its planning algorithm or software system getting stuck or crashing, and so forth, see \tabref{table:incorrect_goal+cause}. 

\begin{figure*}[t!]
\centering
\includegraphics[width=\textwidth]{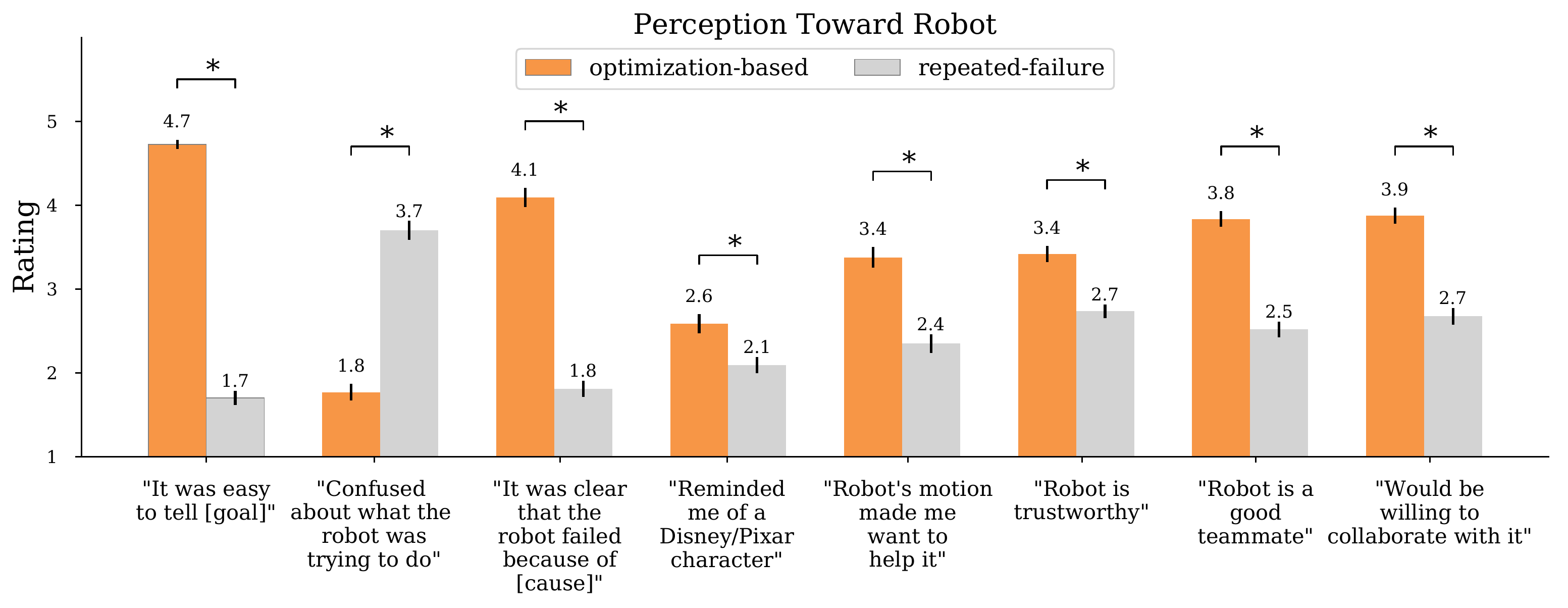}
\caption{Ratings toward the Likert statements used in \sref{sec:mainstudy}. A higher value indicates higher confidence. For each statement, ratings for the optimization-based and repeated-failure conditions were averaged across tasks.}
\label{fig:full_percep}
\end{figure*}

For assessing task goal and cause of incapability, participants saw either the motions expressing incapability or the repeated-failure motions. Next we assessed participants' subjective perceptions and attitudes toward the robot, and for that, we wanted participants to compare the two ``robots''---with expressive motions generated by either our optimization-based approach or the repeated failures. We felt comparisons were important here in order to ground participants' perceptions, thus improving experimental reliability. We thus introduced the other robot and asked them, for each robot, to rate their level of agreement with the statements in \figref{fig:full_percep}.

\begin{figure}[t!]
\centering
\hbox{\hspace{-0.5em}\includegraphics[width=0.5\textwidth]{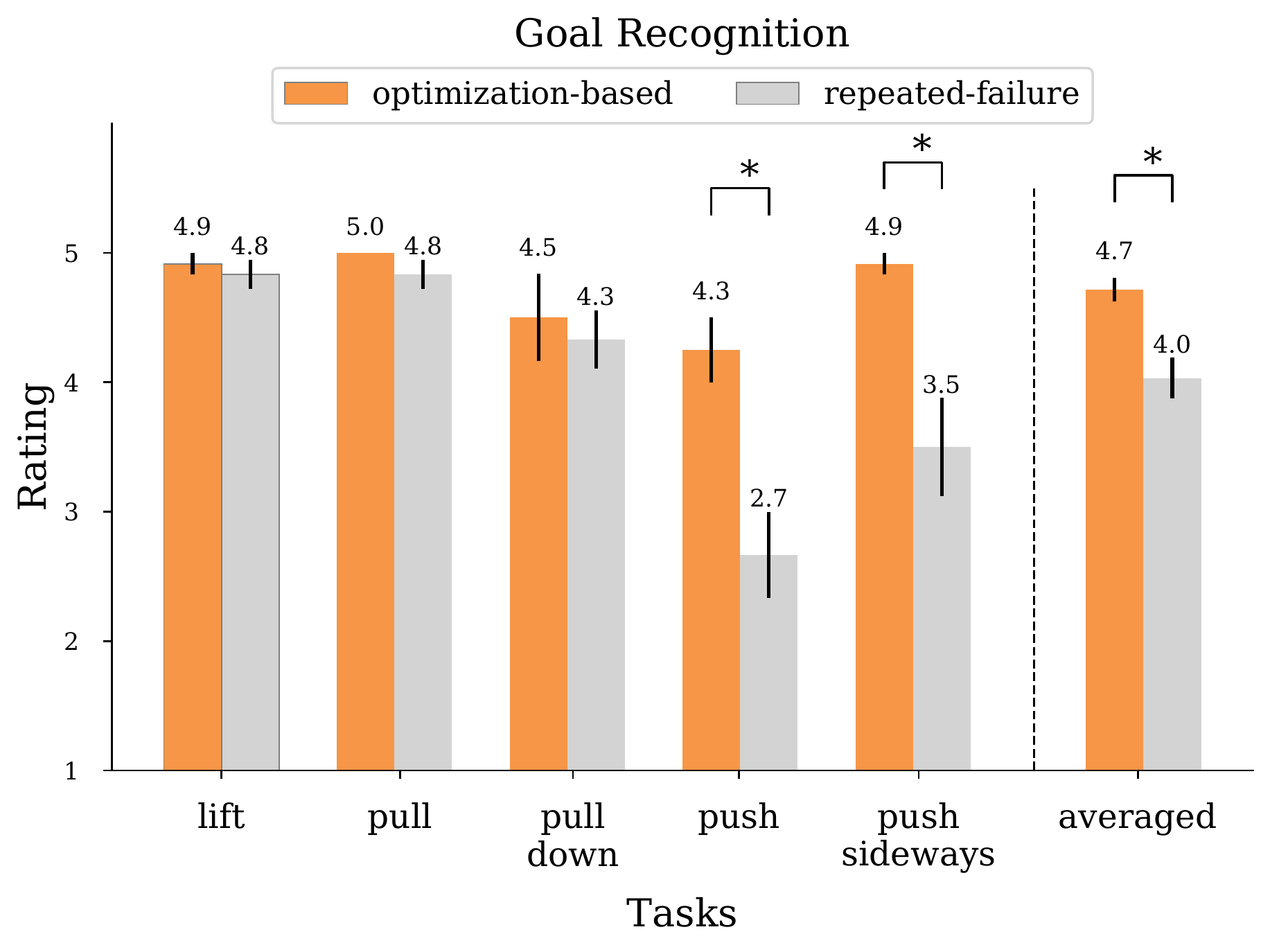}}
\caption{Ratings toward the correct goal for each task. A higher value indicates higher confidence. The averaged values represent mean ratings of expressive and non-expressive motions across tasks.}
\label{fig:goal}
\end{figure}

\prg{Subject Allocation}
We recruited 120 participants (38\% female, median age 33) through Amazon Mechanical Turk. All participants were from the United States and had a minimum approval rating of 95\%. The optimization-based versus repeated-failure manipulation was between-subjects for the first part of the study and was within-subjects for the last part, in which we evaluated subjective perceptions of the robot. We had 24 participants for each of the five tasks, where 12 were in the optimization-based expressive motion condition and the other 12 were in the repeated-failure condition.

\begin{figure}[t!]
\centering
\hbox{\hspace{-0.5em}\includegraphics[width=0.5\textwidth]{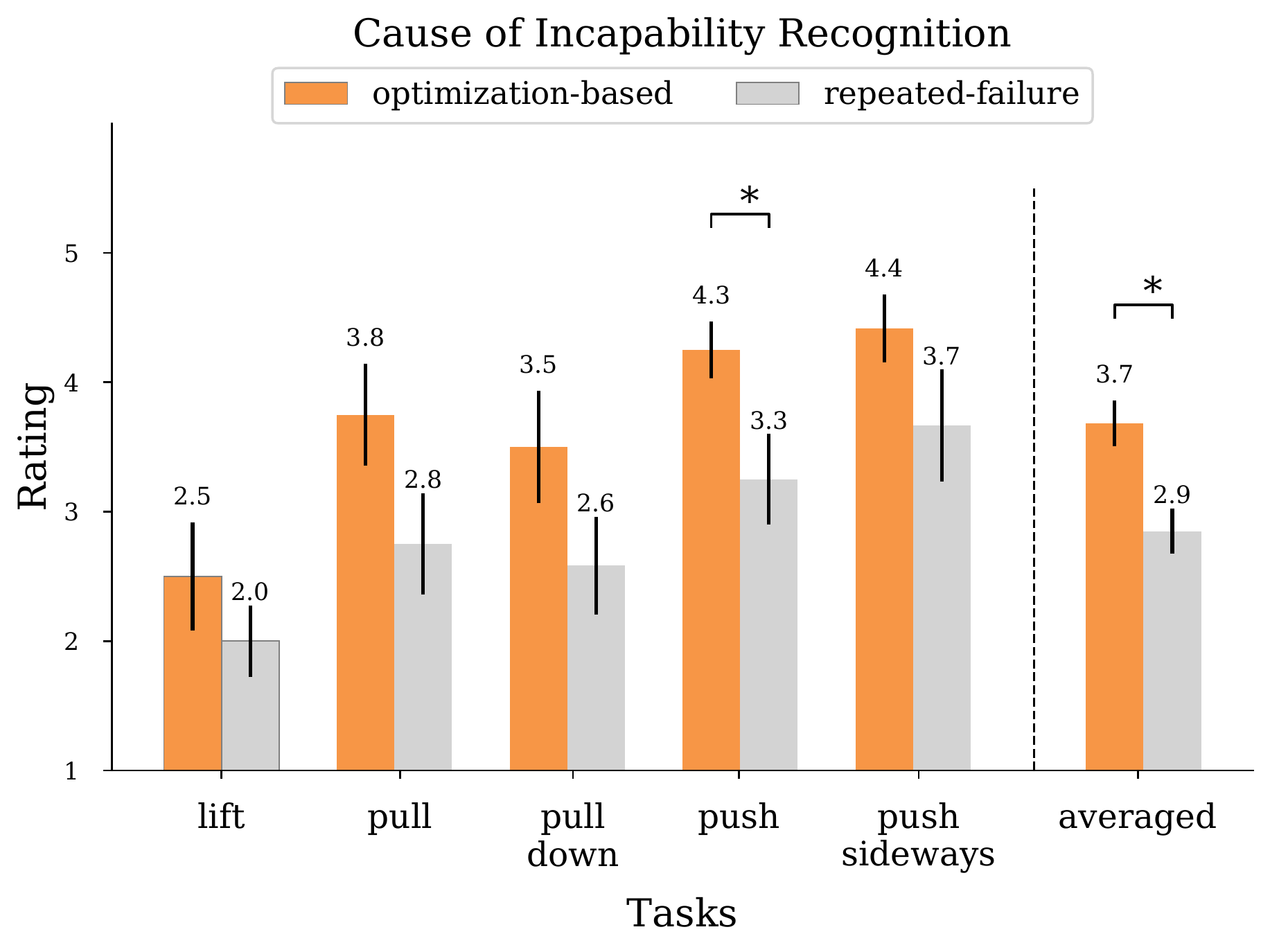}}
\caption{Ratings toward the correct cause of incapability for each task. A higher value indicates higher confidence. The averaged values represent mean ratings of optimization-based and repeated-failure motions across tasks.} 
\label{fig:cause}
\end{figure}

\prg{Hypotheses}
We hypothesized that motions expressing incapability will help participants understand the robot's goal and incapability better than repeated-failures will, across all tasks. We also hypothesized that participants will perceive the robot with motions expressing incapability more positively than they perceive the robot with repeated-failures.

\noindent\textbf{H1}: Motions expressing incapability improve goal recognition.

\noindent\textbf{H2}: Motions expressing incapability improve cause of incapability recognition.

\noindent\textbf{H3}: Participants perceive the robot more positively when it uses motions expressing incapability on an incompletable task.

\subsection{Analysis}

\prg{Goal Recognition}
We first analyzed participants' ratings of the different possible goals. We ran a two-way ANOVA with motion type as the independent variable for each possible goal's rating. We found that motions expressing incapability  significantly improved the rating of the correct goal ($F(1,119)=19.43$, $p<.0001$), and significantly decreased the average rating given to the incorrect goals ($F(1,119)=23.79$, $p<.0001$). This supports our hypothesis \textbf{H1}.

\figref{fig:goal} shows a task breakdown of the correct goal rating. We can see that the largest improvements are in \emph{push} and \emph{push sideways}, probably because the context is not enough for these tasks to be conveyed by the non-expressive motion---some attempt is really needed to understand what the robot is trying to do. In contrast, as the robot reaches for the cup, its intended goal of lifting the cup becomes pretty clear even without an attempt.

Next, we analyzed participant's rankings of the possible goals. Motions expressing incapability significantly improved the ranking of the correct goal ($F(1,119)=30.69$, $p<.0001$) from an average ranking of 1.65 (already close to the top) to one of 1.05, with nearly all participants selecting the correct goal (95\% as opposed to 60\%).

We also analyzed participants' open-ended responses. We categorized an open-ended response as correct if the response contained all keywords, or synonyms of keywords, from the correct goal recognition statement. For example, in the correct statement, ``The robot was trying to pick up the cup,'' we designated ``pick up'' and ``cup'' as keywords. We had two experimenters code the statements where one coder coded all statements and the other coded 10\%. There was strong agreement between the two coders' judgments, with Cohen's $\kappa=1$, $p=.001$. We found that participants who saw the motion expressing incapability were able to explain the robot's goal correctly (in their open-ended response) significantly more than participants who saw the repeated-failure, $p=.0003$.

Finally, we analyzed the two goal-recognition Likert-scale subjective questions at the end of the study (``It was easier to tell [the robot's goal].'' and ``I was confused about what the robot was trying to do.''). We used a repeated-measures ANOVA, since this part was within-subjects. We found that motions expressing incapability improved the ease of identifying the goal ($F(1,119)=602.38$, $p<.0001$) and decreased confusion about the goal ($F(1,119)=183.13$, $p<.0001$).

\begin{quote}
  \emph{Overall, our results support \textbf{H1}: our motions expressing incapability improved goal recognition.}  
\end{quote}

\prg{Cause of Incapability Recognition}
Looking first at the ratings for possible causes of incapability, participants rated five causes, where one was the robot's actual cause of incapability, and another was that the robot was not strong enough---a plausible cause that is hard to disambiguate from the actual cause within each task. The other causes were incorrect. We found that motions expressing incapability significantly improved the rating of the correct cause ($F(1,119)=13.6,$ $p=.0003$), but did not have a significant effect on the rating of the plausible cause. Motions expressing incapability decreased the average ratings of the incorrect causes ($F(1,114)=19.05$, $p<.0001$). 

Motions expressing incapability also significantly improved the rank of the correct cause ($F(1,119)=23.71$, $p<.0001$), from an average of 3.02 to an average of 1.98.  

Next, we looked at participants' open-ended responses. We used the same coding scheme as we did for the goal-recognition open-ended statements.
We found that there was a  strong agreement between the two coders' judgments, with Cohen's $\kappa=1$, $p = .001$.
Participants who saw the motions expressing incapability were significantly more likely to describe the correct cause of incapability compared to those who saw the repeated-failures, $p=.0002$. 

Finally, we conducted a repeated-measures ANOVA on the one subjective rating relevant to cause of incapability recognition (``It was clear that [cause of incapability].''). We found motions expressing incapability significantly improved this rating  ($F(1,119)=182.31$, $p<.0001$). 

\begin{quote}
\emph{Overall, our results support \textbf{H2}: our method for generating motions expressing incapability improved cause of incapability recognition.}
\end{quote}

\prg{Perception of Robot}
We ran a repeated-measures ANOVA for each statement. With motions expressing incapability, users perceived the robot as more like an animated character ($F(1,119)=30.82$, $p<.0001$), wanted to help the robot more ($F(1,119)=51.93$, $p<.0001$), thought it was more trustworthy ($F(1,119)=31.76$, $p<.0001$) and a better teammate ($F(1,119)=85.97$, $p<.0001$), and were more willing to collaborate with the robot in the future ($F(1,119)=69.66$, $p<.0001$). See \figref{fig:full_percep} for details.

\begin{quote}
\emph{Overall, our results support \textbf{H3}: our method for generating motions expressing incapability improved users' perceptions of the robot.}
\end{quote}

\section{Discussion}
\noindent\textbf{Summary.}
We use an optimization-based approach to automatically generate expressive trajectories that communicate \emph{what} a robot is trying to do and \emph{why} it will fail. The optimization produces a trajectory where body points on the robot ``mimic'' how the end-effector would move if the robot had been capable of completing the task. We complemented the expressiveness of our optimized trajectory by manipulating repetition and timing. Our results show that, compared to the state-of-the-art approach, motions expressing incapability improve intent recognition and cause of incapability inference while also increasing positive evaluations of the robot.

Our optimization enables robots to automatically and efficiently generate motions expressing incapability. On average, solving \eref{eqn:prob_form} for for cost $c_{\text{ee}}$, distance $d_{\text{proj}}$, and a specified $q_f$ takes half a second when the base does not move (e.g., for the \emph{lift} and \emph{pull down} tasks) and a few seconds when the base is able to move. Given this, it is feasible for the robot to detect incapability in the middle of execution, and compute an expressive motion on the fly.

\noindent\textbf{Limitations and Future Work.}
Perhaps the greatest limitation of our work is that our optimization covers only a narrow set of tasks the robot is incapable of completing. Incapabilities that are not about the end-effector position changing, such as grasping, or incapabilities that have nothing to do with the end-effector, such as those related to perception, cannot be communicated using our optimization.
Our approach also does not address when or how expressive motions should be accompanied by other channels of communication, such as verbal communication. Although these are very important areas for future work, we are excited to see that the same method can \emph{automatically} generate motion that is expressive and useful \emph{across a range} of different tasks.

\section{Acknowledgments}
This research was funded in part by the Air Force Office of Scientific Research and by HKUST ITF/319/16FP. Sandy Huang was supported by an NSF Fellowship.

\bibliographystyle{ACM-Reference-Format}
\bibliography{references} 

\end{document}